%% file: root.tex
\documentclass[letterpaper, 10 pt, conference]{ieeeconf}  

\IEEEoverridecommandlockouts                              

\usepackage{url}
\usepackage{graphicx}
\usepackage{tikz}
\usepackage{xcolor}
\usepackage{censor}

\title{\LARGE \bf
Agentic LLM Planning via Step-Wise PDDL Simulation: An Empirical Characterisation
}

\author{Kai G\"obel, Pierrick Lorang, Patrik Zips, and Tobias Gl\"uck%
\thanks{All authors are with AIT Austrian Institute of Technology GmbH, Vienna, Austria
        (e-mail: {\tt\small \{kai.goebel, pierrick.lorang, patrik.zips, tobias.glueck\}@ait.ac.at})}%
\thanks{PyPDDLEngine is available at \protect\url{https://github.com/AIT-Complex-Dynamical-Systems/pypddlengine}}%
}

\begin{document}

\maketitle
\thispagestyle{empty}
\pagestyle{empty}

\input{sections/0_abstract}

\input{sections/1_introduction}
\input{sections/2_related_work}

\input{sections/3_methods}
\input{sections/4_results}
\input{sections/5_discussion}
\input{sections/6_conclusion}




\section*{ACKNOWLEDGMENT}

The authors used Claude Sonnet 4.6 (Anthropic) as a writing assistant in drafting and editing the manuscript across all sections. After using this tool, the authors reviewed and edited the content as needed and take full responsibility for the content of the publication.

\bibliographystyle{IEEEtran}
\bibliography{sections/mybib}

\end{document}

%% file: sections/0_abstract.tex
\begin{abstract}
Task planning, the problem of sequencing actions to reach a goal from an initial state, is a core capability requirement for autonomous robotic systems.
Whether large language models (LLMs) can serve as viable planners alongside classical symbolic methods remains an open question.
We present PyPDDLEngine, an open-source Planning Domain Definition Language (PDDL) simulation engine that exposes planning operations as LLM tool calls through a Model Context Protocol (MCP) interface.
Rather than committing to a complete action sequence upfront, the LLM acts as an interactive search policy that selects one action at a time, observes each resulting state, and can reset and retry.
We evaluate four approaches on 102 International Planning Competition (IPC) Blocksworld instances under a uniform 180-second budget: Fast Downward lama-first and seq-sat-lama-2011 as classical baselines, direct LLM planning (Claude Haiku 4.5), and agentic LLM planning via PyPDDLEngine.
Fast Downward achieves 85.3\% success. The direct and agentic LLM approaches achieve 63.7\% and 66.7\%, respectively, a consistent but modest three-percentage-point advantage for the agentic approach at $5.7\times$ higher token cost per solution.
Across most co-solved difficulty blocks, both LLM approaches produce shorter plans than seq-sat-lama-2011 despite its iterative quality improvement, a result consistent with training-data recall rather than generalisable planning.
These results suggest that agentic gains depend on the nature of environmental feedback. Coding agents benefit from externally grounded signals such as compiler errors and test failures, whereas PDDL step feedback is self-assessed, leaving the agent to evaluate its own progress without external verification.
\end{abstract}

%% file: sections/1_introduction.tex
\section{Introduction}

\begin{figure*}[!t]
  \centering
  \includegraphics[width=0.85\textwidth]{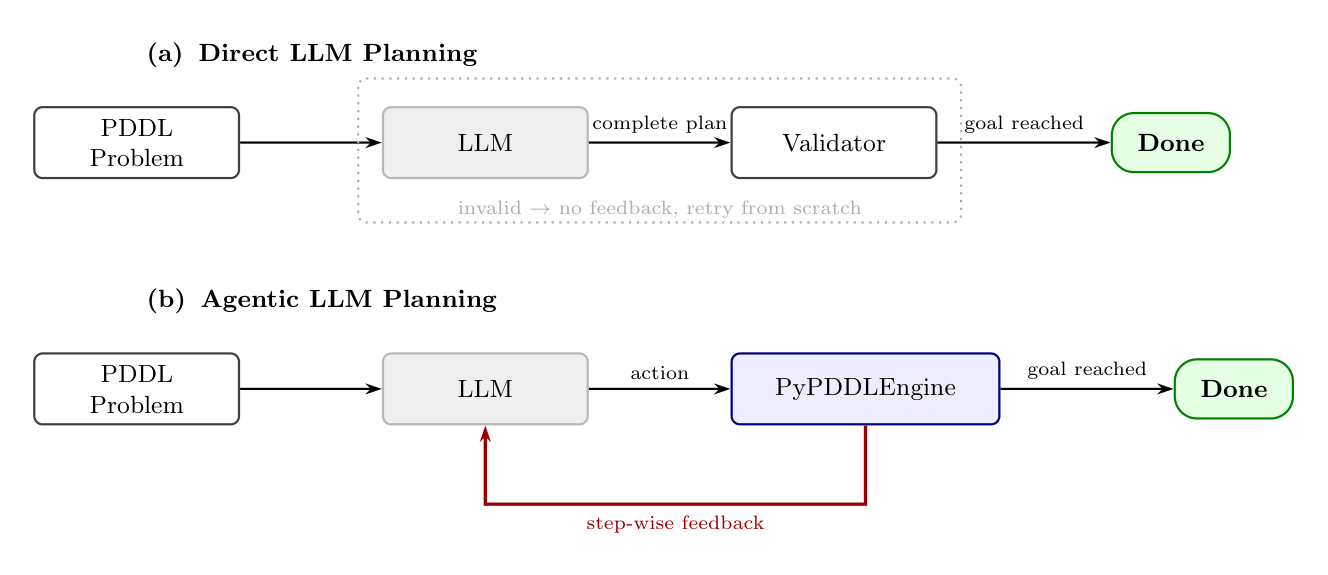}
  \caption{The two LLM planning approaches evaluated in this work.
    \textit{(a)} Direct LLM planning generates a complete plan in a single
    pass. An invalid result is discarded and the model is re-prompted from
    scratch with no feedback between attempts.
    \textit{(b)} Agentic LLM planning executes one action at a time, with
    PyPDDLEngine providing step-wise state feedback after each action,
    closing the loop between the LLM and the planning environment.}
  \label{fig:open_closed_loop}
\end{figure*}

Autonomous robotic systems require planners that can sequence actions from an initial state to a desired goal configuration, a capability known as task planning.
Classical planners solve these problems efficiently through systematic symbolic search over a formal domain model.
Large language models (LLMs) take a fundamentally different approach, drawing on broad world knowledge and common-sense reasoning acquired from internet-scale training to generate structured action sequences from natural-language or symbolic task descriptions, without domain-specific programming.
Whether LLMs can serve as viable alternatives or complements to classical planners, and under what conditions each approach succeeds or fails, is a concrete open question for robotics~\cite{valmeekam2023planning,kambhampati2024llms}.

Agentic LLM systems interact iteratively with an environment, observe feedback after each action, and adjust their behaviour accordingly~\cite{chen2023llmstate,wang2023voyager}.
In software engineering, coding agents that execute code, observe failures, and revise their solutions achieve top performance on real-world programming benchmarks~\cite{shinn2023reflexion,yang2024sweagent}.
This success has fuelled speculation that the same interactive paradigm could unlock stronger performance across other reasoning domains, including task planning.

Existing work on LLM planning focuses on direct generation, prompting strategies, and re-sampling with additional context~\cite{valmeekam2023planning,kambhampati2024llms,stechly2024self}.
Running an LLM as an agent inside a Planning Domain Definition Language (PDDL) simulation engine, where it executes one action at a time, observes the resulting world state, and can reset and retry when a trajectory is judged unproductive, has received limited systematic study.
This is architecturally distinct from online replanning, in which a classical planner is re-invoked after each real-world action from the newly observed state. Here the LLM itself acts as the search policy, navigating the PDDL state space within a simulation rather than delegating search to a classical planner.
Classical planners and PDDL validators do not support interactive step-wise decision-making, leaving no tool that allows an LLM to execute one action, observe the resulting state, and decide the next action on that basis.
Whether step-wise simulation feedback improves LLM planning performance, and under what conditions, remains an open question.

We present PyPDDLEngine, a Python library implementing a PDDL simulation engine and exposing its planning operations as LLM tool calls through a Model Context Protocol (MCP) interface, enabling genuine step-wise interaction with a live planning environment.
PyPDDLEngine supports seven operations: initialise a problem, query the current state, retrieve applicable actions, execute a single action, reset to the initial state, review action history, and validate a complete plan.
Unlike PDDL validators that accept or reject a finished plan, PyPDDLEngine allows the LLM to observe each state transition as it plans.
This positions the LLM as the search policy. Rather than generating a plan upfront or consuming a classically computed sequence, it navigates the PDDL state space one action at a time, with the transition model provided by the engine.
We release PyPDDLEngine as open-source software to support reproducible agentic planning research.

We compare four approaches on 102 International Planning Competition (IPC) Blocksworld instances under a uniform 180-second time budget: direct LLM planning (Claude Haiku 4.5), agentic LLM planning via PyPDDLEngine, Fast Downward~\cite{helmert2006fast} lama-first (satisficing, greedy), and Fast Downward seq-sat-lama-2011 (satisficing, anytime).
Blocksworld provides a well-defined difficulty gradient across 102 instances and is widely used in prior LLM planning research, making results directly comparable~\cite{valmeekam2023planning,stechly2024self,valmeekam2024llms}.
We measure success rate, plan length on co-solved instances, failure modes, and token cost.

Our main contributions are:
\begin{itemize}
  \item \textbf{PyPDDLEngine}: an open-source PDDL simulation engine with an MCP interface, positioning the LLM as an interactive search policy within the PDDL transition system rather than as a plan generator or translator, and released to support reproducible agentic planning research.
  \item \textbf{Empirical characterisation}: a four-way evaluation of direct LLM, agentic LLM, and two classical planning baselines across 102 Blocksworld instances, quantifying the scope, cost, and difficulty-dependence of the agentic advantage, and showing that the modest gains are consistent with the self-assessed nature of PDDL step feedback, in contrast to the externally grounded signals that drive large gains in coding agents.
\end{itemize}

%% file: sections/2_related_work.tex
\section{Related Work}

Systematic benchmarks establish that LLMs fail to reliably generate executable plans in symbolic planning domains.
Valmeekam et al.\ document this failure on IPC benchmarks and provide a key diagnostic~\cite{valmeekam2023planning,valmeekam2024llms}. When action names are syntactically relabelled, success rates collapse to near zero, pointing to approximate retrieval from training data rather than genuine reasoning~\cite{kambhampati2024llms}.
Asking the LLM to critique its own unexecuted plan text does not improve performance, and Stechly et al.\ demonstrate that gains from iterative prompting trace to repeated sampling under an external verifier rather than to the critique itself~\cite{stechly2024self}.
Whether an LLM can improve by reflecting on the outcomes of \emph{executed} actions, observing real state transitions rather than its own plan text, is a distinct question that this prior work does not directly address.
Reasoning-focused models improve accuracy on small instances but remain brittle on harder problems and carry substantially higher inference costs than classical planners~\cite{valmeekam2024llms}.
Broader surveys consolidate the landscape~\cite{pallagani2024prospects,tantakoun2025llms,armony2025how}.

A dominant architecture in LLM-PDDL research positions the model as a translator, where the LLM generates a PDDL problem or domain description that is then passed to a classical solver, delegating correctness to the downstream planner~\cite{liu2023llmp,silver2023generalized,guan2023leveraging,mahdavi2024leveraging,hao2024planning}.
Variants of this architecture address automated domain acquisition~\cite{smirnov2024generating,babu2025adaptive,huang2024planning,liu2024delta}, end-to-end agentic plan formulation~\cite{lamalfa2025endtoend}, combined task-and-motion planning~\cite{chen2023autotamp,chu2025llmmap}, action language reasoning~\cite{ishay2025llmal}, and search heuristic generation for classical planners~\cite{correa2025classical}.
In all of these approaches, PDDL is a text artefact the LLM generates for a downstream solver to consume.
Our work takes a different position: rather than placing the LLM upstream of the solver, we embed it inside a live PDDL simulation, where it executes actions one at a time and observes each resulting state before choosing the next action.

ReAct established the paradigm of interleaving reasoning traces and environment-grounded actions in a closed loop, demonstrating that this pattern reduces hallucination and improves decision-making~\cite{yao2022react}.
Subsequent work applied this pattern to embodied and robotic planning, with approaches ranging from LLMs that generate step-by-step plans with replanning on failure~\cite{song2022llmplanner}, to code-inspired prompts with executable assert statements~\cite{singh2022progprompt}, to iterative self-refinement and feedback integration in robot task execution~\cite{zhou2023isrllm,ding2023integrating,zhao2023large,kwon2024fast}.
In open-world settings, closed-loop operation requires implicit state reconstruction through language inference because no formal transition model is available~\cite{chen2023llmstate}.
Embodied exploration agents extend this paradigm further to open-ended environments without predefined task structures~\cite{wang2023voyager}.
Recent surveys introduce a structural taxonomy for the field, distinguishing LLM-as-Planner systems, which reason without a formal transition model, from LLM-as-Facilitator systems, which translate problems for a classical planner~\cite{li2024lasp}.
Search-based approaches that embed the LLM as a search policy over a formal world model represent a structurally distinct third position~\cite{wei2025plangenllms}.
Our work instantiates this third position: PyPDDLEngine provides the transition model and the expansion oracle while the LLM acts as the search policy, embedding the model inside the search loop rather than upstream of it.
We also assess empirically whether this grounded, step-wise interaction produces meaningful gains over direct generation, a question prior work has not directly addressed.

%% file: sections/3_methods.tex
\section{Methods}

\subsection{PyPDDLEngine and MCP Interface}

Classical PDDL planners such as Fast Downward~\cite{helmert2006fast} solve problems end-to-end, committing to a complete action sequence through internal search without supporting interactive step-wise decision-making during plan construction. PDDL validators such as VAL accept or reject complete action sequences but do not support interaction during plan construction.
Neither tool class allows an LLM agent to execute one action, observe the resulting state, and decide the next action on that basis.
PyPDDLEngine is a Python library that fills this infrastructure gap. It wraps a PDDL simulation engine and exposes the PDDL transition system as an interactive environment, enabling step-wise LLM interaction with a live planning environment.
We release PyPDDLEngine as open-source software to support reproducibility.

PyPDDLEngine exposes seven operations:
\begin{itemize}
  \item \texttt{initialise\_session}: loads domain and problem files or strings and resets the environment to the initial state.
  \item \texttt{query\_current\_state}: returns the set of currently true predicates.
  \item \texttt{query\_applicable\_actions}: returns all ground actions valid in the current state.
  \item \texttt{execute\_single\_action}: applies one action and advances the state, returning a success flag, the updated state, and a goal-reached flag.
  \item \texttt{reset\_to\_initial\_state}: restores the environment to the problem's initial state.
  \item \texttt{query\_action\_history}: returns the sequence of actions executed in the current session.
  \item \texttt{validate\_complete\_plan}: checks whether a submitted action sequence satisfies all goal conditions.
\end{itemize}
The engine supports STRIPS-style PDDL with typing. Derived predicates, numeric fluents, and durative actions are not supported.

PyPDDLEngine is wrapped as an MCP server, providing a standardised JSON-RPC interface through which LLMs invoke tools by name.
Each of the seven operations is registered as an MCP tool with typed input and output schemas, and the agentic LLM receives these tool descriptions at context initialisation.
The MCP interface decouples the engine from any specific LLM provider, enabling experiments across models without code changes.

\subsection{Planning Approaches}

The direct approach provides Claude Haiku 4.5 (temperature 0.2) with the full PDDL domain and problem description as text in a single prompt.
Temperature 0.2 is chosen to introduce stochastic variation across independent retry attempts, so that successive runs can produce different candidate plans.
The model returns a complete action sequence, which is independently validated. If the sequence is invalid, the context is fully reset and the model is re-prompted from scratch with no failure feedback.
Retry attempts continue until a valid plan is found or the 180-second wall-clock budget is exhausted.
The direct approach does not use PyPDDLEngine and no step-wise interaction occurs (Fig.~\ref{fig:open_closed_loop}a).
We use Claude Haiku 4.5 for a budget-controlled study design, not as a claim about frontier model capability, since frontier reasoning models carry substantially higher inference costs~\cite{valmeekam2024llms}.

The agentic approach provides Claude Haiku 4.5 with the PDDL domain, the problem description, and the full set of PyPDDLEngine MCP tool descriptions.
The model executes actions one at a time using \texttt{execute\_single\_action} and observes each resulting state before choosing the next action.
Unlike online replanning, the entire process occurs within the PDDL simulation, with the LLM acting as the search policy at each step rather than following a pre-computed sequence.
\texttt{reset\_to\_initial\_state} is available to restart from scratch when the model judges the current trajectory unproductive, and we observe this behaviour in practice.
System prompts for both approaches were iteratively refined through trace analysis to elicit reliable behaviour, though prompt design is not the focus of this study.
Interaction continues until \texttt{validate\_complete\_plan} confirms a valid plan or the 180-second wall-clock budget is exhausted (Fig.~\ref{fig:open_closed_loop}b).
Temperature is not configurable in the agentic framework.

Two Fast Downward~\cite{helmert2006fast} configurations serve as classical planning baselines.
Fast Downward lama-first (satisficing, greedy) finds a valid plan as quickly as possible without iterative improvement. It is the primary classical baseline for comparison with the LLM approaches because, like them, it commits to a single plan without post-hoc refinement.
Fast Downward seq-sat-lama-2011 (satisficing, anytime) begins with a greedy solution and iteratively shortens it within the time budget. We include it as a plan-quality reference rather than a direct comparator, to provide a near-optimised length baseline against which LLM plan lengths can be assessed.
Both configurations run under the same 180-second wall-clock budget and never produce an invalid plan.

\subsection{Experimental Protocol}

We evaluate all four approaches on 102 Blocksworld instances from the IPC, used in their original benchmark order.
We select Blocksworld because it provides a smooth, well-defined difficulty gradient and is systematically covered in prior LLM planning research, making results directly comparable~\cite{valmeekam2023planning,stechly2024self,valmeekam2024llms}.
Fast Downward lama-first plan length serves as the primary difficulty proxy, preferred over block count alone because complexity depends on both the number of blocks and the goal configuration. A problem with many blocks but a trivial goal may still require only a few steps.

We measure four quantities for each approach:
\begin{itemize}
  \item \textbf{Success rate}: the fraction of instances for which a valid, goal-satisfying plan is returned within the time budget, confirmed by independent plan validation rather than process termination alone.
  \item \textbf{Plan length}: the number of actions in the returned plan, recorded only for successful runs.
  \item \textbf{Failure mode}: one of three categories -- solved (goal reached), timed out (180 seconds elapsed), or early exit (an agentic-approach-specific outcome in which the model judges the problem unsolvable and halts before the budget expires).
  \item \textbf{Token cost}: total input and output tokens consumed per run, converted to tokens per solved instance for the LLM approaches.
\end{itemize}

All four approaches share the same 180-second wall-clock time budget.
All experiments were conducted on a workstation running Ubuntu 24.04.4 LTS with an Intel Core i9-14900K (24 cores, 48 threads, up to 6.0\,GHz) and 128\,GB RAM. Fast Downward wall-clock times are hardware-relative and should be interpreted in this context.
Plan quality is compared only on the subset of instances solved by all four approaches, referred to as the co-solved set, to avoid survivorship bias. An approach that solves only easy instances would otherwise appear to produce shorter plans not because it plans more efficiently, but because it never attempts the harder instances that require longer solutions.

Each approach is evaluated once per instance. Running multiple trials to estimate variance was not feasible given token cost.

%% file: sections/4_results.tex
\section{Results}

\definecolor{cbblue}{RGB}{0,114,178}   
\definecolor{cborange}{RGB}{230,159,0} 
\definecolor{cbgray}{RGB}{153,153,153} 


\subsection{Benchmark Difficulty Structure}

\begin{table}[h]
\caption{Mean plan length (actions) and instance count by difficulty block for Fast Downward. Both configurations solve identical instance sets. $n$ is the number of instances solved in each block. $\Delta$ = lama-first mean $-$ seq-sat-lama-2011 mean. Block 100--102 contains two instances, both unsolved within budget.}
\label{tab:fd_plan_length}
\centering
\begin{tabular}{lrrrr}
\hline
Block    & $n$ & FD\,lama & FD\,seq & $\Delta$ \\
\hline
0--10    & 10  &  14.0 &  14.0 &    0.0 \\
10--20   & 10  &  39.8 &  37.6 &    2.2 \\
20--30   & 10  &  61.6 &  36.4 &   25.2 \\
30--40   & 10  & 146.4 &  64.8 &   81.6 \\
40--50   & 10  & 199.2 & 106.8 &   92.4 \\
50--60   & 10  & 222.4 & 135.4 &   87.0 \\
60--70   & 10  & 253.8 & 207.2 &   46.6 \\
70--80   &  7  & 281.4 & 228.3 &   53.1 \\
80--90   &  7  & 380.0 & 328.0 &   52.0 \\
90--100  &  3  & 484.0 & 374.7 &  109.3 \\
100--102 &  0  & ---   & ---   &   ---  \\
\hline
\end{tabular}
\end{table}

Fast Downward lama-first plan length grows monotonically across difficulty blocks (Table~\ref{tab:fd_plan_length}), rising from 14 actions at blocks 0--10 to 484 actions at blocks 90--100 across the range of instances both configurations can solve within budget.
Fast Downward seq-sat-lama-2011, which iteratively shortens its solution within the time budget, produces plans of equal length to lama-first at blocks 0--10 and shorter plans at all higher difficulty levels.
The mean reduction widens from 2.2 actions at blocks 10--20 to 109.3 actions at blocks 90--100.
Together, the two configurations bracket the plausible plan-length range for each instance, with lama-first establishing a greedy upper bound and seq-sat-lama-2011 a near-optimised lower bound.

Both Fast Downward approaches time out on the same 15 instances (instance 72 onwards in the ordered set), confirming a shared difficulty ceiling under the 180-second budget.

\subsection{Success Rate and Failure Modes}

Table~\ref{tab:results} summarises success rates and failure modes. Both Fast Downward configurations solve identical instance sets, with all failures attributable to timeouts on the same 15 hard instances. The agentic approach achieves a three percentage point advantage over the direct approach.

\begin{table}[h]
\caption{Success rates and failure modes across 102 IPC Blocksworld instances.}
\label{tab:results}
\centering
\begin{tabular}{lrrr}
\hline
Approach & Solved & Timeout & Early exit \\
\hline
FD lama-first    & 87 (85.3\%) & 15 & 0 \\
FD seq-sat-lama-2011 & 87 (85.3\%) & 15 & 0 \\
Direct LLM       & 65 (63.7\%) & 37 & 0 \\
Agentic LLM      & 68 (66.7\%) & 28 & 6 \\
\hline
\end{tabular}
\end{table}

The direct approach produces two possible outcomes, either a valid plan within budget or a budget timeout. On all unsolved instances, the budget expires without a valid plan found.
The agentic approach introduces a third outcome absent from all other configurations. On six instances (32, 88, 89, 96, 98, 100), the model judges the problem unsolvable and halts before the budget expires, producing an early exit.
On four of these six instances (32, 88, 89, and 100), the direct approach finds a valid plan, demonstrating that the early exit represents an incorrect assessment of unsolvability rather than a recognition of genuine difficulty. On the remaining two instances (96 and 98), both LLM approaches fail.

\begin{figure*}[t]
  \centering
  \resizebox{\textwidth}{!}{\includegraphics[width=0.95\textwidth]{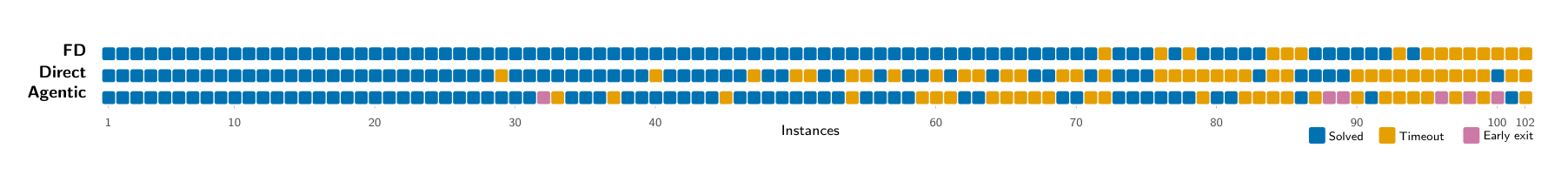}}
  \caption{Per-instance outcomes for all 102 IPC Blocksworld instances, ordered by index (difficulty generally increases left to right). Both LLM approaches begin failing earlier, with the agentic approach showing six early exits where the model incorrectly judged the problem unsolvable before the budget expired.}
  \label{fig:per_instance}
\end{figure*}

Fig.~\ref{fig:per_instance} shows per-instance outcomes across all 102 instances. Both Fast Downward approaches maintain 100\% success through blocks 60--70. The LLM approaches begin declining from blocks 20--30 onwards. The agentic approach tracks slightly above the direct approach through the mid-difficulty range. In blocks 80--90, this advantage inverts. The agentic approach achieves 20\% success against the direct approach's 50\%, a pattern not observed in adjacent difficulty blocks.

The direct approach generates a complete candidate plan per attempt and retries on validation failure. On instances it solves, the median number of attempts is 2, indicating that success is reached quickly. On instances it fails, the median is 15, confirming that budget exhaustion results from repeated unsuccessful sampling rather than a single slow generation.

\subsection{Token Cost}

The direct approach consumes a mean of 28,488 tokens per run, against 169,864 tokens for the agentic approach, a ratio of $5.97\times$.
Normalised to solved instances, the direct approach costs 44,705 tokens per solution and the agentic approach costs 254,796, a ratio of $5.7\times$.
The three additional solutions produced by the agentic approach over the direct approach correspond to approximately 14.4 million additional tokens in total.

\subsection{Plan Quality on Co-Solved Instances}

\begin{table}[h]
\caption{Mean plan length (actions) on co-solved instances by block. $n$ is the number of instances solved by all four approaches. Blocks with no co-solved instances are omitted. FD\,lama = Fast Downward lama-first, FD\,seq = Fast Downward seq-sat-lama-2011. Bold marks the shortest plan length per row.}
\label{tab:plan_length}
\centering
\begin{tabular}{lrrrrr}
\hline
Block  & $n$ & FD\,lama & FD\,seq & Direct & Agentic \\
\hline
0--10  & 10  & \textbf{14.0}  & \textbf{14.0}  & \textbf{14.0}  & 15.8  \\
10--20 & 10  & 39.8  & 37.6  & \textbf{27.4}  & 30.6  \\
20--30 &  9  & 60.0  & \textbf{36.2}  & 40.4  & 44.7  \\
30--40 &  6  & 164.3 & 63.3  & \textbf{59.7}  & 60.3  \\
40--50 &  7  & 197.1 & 92.6  & \textbf{78.0}  & 78.6  \\
50--60 &  4  & 215.5 & 144.5 & \textbf{99.5}  & 124.5 \\
70--80 &  3  & 142.0 & \textbf{136.7} & 139.3 & 139.3 \\
\hline
\end{tabular}
\end{table}

Plan quality comparisons are restricted to the 49 instances solved by all four approaches, referred to as the co-solved set, to prevent survivorship bias. Table~\ref{tab:plan_length} reports mean plan lengths by block.
Across most co-solved blocks, both LLM approaches produce shorter plans than seq-sat-lama-2011, despite that configuration actively iterating toward shorter solutions within the budget. Fast Downward lama-first produces the longest plans in all blocks, as expected from greedy planner design.
The one exception is blocks 20--30, where seq-sat-lama-2011 falls below both LLM approaches. At blocks 70--80 only three co-solved instances remain and all four approaches produce nearly identical plan lengths. No co-solved instances exist at higher difficulty levels.

\subsection{Hard-Case Analysis}

The 15 instances on which both Fast Downward approaches time out constitute the hardest subset of the benchmark under the 180-second budget.
On these 15 instances, both LLM approaches solve one common instance (instance 86).
The agentic approach alone solves three instances the direct approach does not (instances 76, 78, and 101).
The direct approach alone solves one instance the agentic approach does not (instance 100).
Neither LLM approach solves the remaining 10 of the 15 Fast Downward-timeout instances.

Instance 101 is the most striking single-instance result for the agentic approach. Both Fast Downward configurations and the direct LLM approach time out, while the agentic approach returns a valid 186-action plan with a time-to-solution of 108--136 seconds.
A single instance is insufficient to distinguish reliable capability from a favourable trajectory, and we note it as an outlier that deserves further investigation.

%% file: sections/5_discussion.tex
\section{Discussion}

\subsection{Scope Clarification}

This study does not assess whether LLMs can replace classical planners in production settings.
Our goal is to characterise LLM planning behaviour under controlled experimental conditions, measuring specifically whether agentic step-wise interaction improves success rates compared to direct generation, and identifying what mechanism may explain the observed pattern.
Like all PDDL-based planning research, we evaluate the planning module in isolation, where the domain model and initial state are given inputs and questions of perception, state estimation, and physical execution are outside scope and orthogonal to the planning questions studied here.

\subsection{LLM Plan Length on Co-Solved Instances}

On the 49 co-solved instances, both LLM approaches produce plans shorter than those of seq-sat-lama-2011 across most difficulty blocks.
Fast Downward seq-sat-lama-2011 has a structural advantage that no other approach in this comparison shares, as it actively iterates toward shorter solutions within the time budget.
It does not guarantee optimality, since the quality of its output depends on how many improvement iterations complete within 180 seconds.
That both LLM approaches nonetheless produce shorter plans across most co-solved difficulty blocks is notable. Without any quality-improvement mechanism, they match or beat a planner specifically designed to shorten its solutions.

One partial explanation is anytime convergence, since seq-sat-lama-2011 begins with a greedy solution and iterates with progressively tighter cost bounds, and on harder instances fewer improvement iterations may complete within the time budget, leaving its returned plans further from optimal.
This explanation is incomplete, however, because seq-sat-lama-2011 is specifically designed to improve plan quality over time, and LLMs beating its anytime result across most difficulty blocks requires a stronger account.

A further observation sharpens this picture. The direct and agentic approaches produce plans of nearly identical length on co-solved instances across all difficulty blocks, with the direct approach occasionally shorter.
The agentic approach has access to step-wise state feedback and can replan, yet it confers no plan-quality advantage over a single-shot generation.
The pattern is inconsistent with a model that reasons incrementally toward a better solution. If the interactive loop were driving quality, the agentic approach should produce shorter plans than the direct approach.
The data instead suggest a binary competence profile. When the LLM succeeds, it produces a near-optimal plan regardless of whether it has access to interactive feedback. When it does not have the solution, neither direct generation nor agentic interaction recovers it.
This contrasts with the graduated performance one would expect from a reasoner. A human solver of intermediate skill might produce a valid but suboptimal plan and improve with effort. The LLM data show no such middle ground.

Blocksworld is one of the most extensively documented planning domains in the literature and in online resources, and the LLM approaches may be recalling near-optimal solution patterns from training data rather than deriving plans by search.
This account is consistent with evidence that LLM performance on Blocksworld collapses to near zero when action names are syntactically relabelled~\cite{valmeekam2023planning,valmeekam2024llms}, and with the binary competence profile observed here.
It cannot be confirmed without controlled experiments varying domain familiarity.

Taken together, the plan quality results suggest that LLM performance on Blocksworld may reflect training-distribution recall rather than generalisable planning.

\subsection{Step-Wise Feedback Provides No Global Progress Signal}

Agentic systems applied to software engineering have achieved strong results on real-world programming benchmarks~\cite{shinn2023reflexion,yang2024sweagent}, yet the agentic approach here outperforms direct generation by only three percentage points.
This contrast requires explanation.

In software engineering, execution feedback is externally grounded. A failing test or a runtime error is produced by the environment, not by the model itself.
Reflexion demonstrates this concretely, showing that agents guided by test-runner feedback achieve large performance gains through verbal reinforcement without weight updates~\cite{shinn2023reflexion}.
The feedback is directional and objective. It identifies the specific location and nature of the fault independently of the model's own assessment, and cannot be misinterpreted.

In step-wise PDDL interaction, the agent has a genuine opportunity for self-directed correction. After each action it observes the updated state, can reflect on its action history, and can reset to the initial state if it judges the current trajectory unproductive, behaviour we observe in practice.
The feedback it acts on is, however, self-generated.
The per-step confirmation tells the agent only that an action was applicable. It does not indicate how far the agent is from the goal or whether the current trajectory is productive.
Progress assessment is therefore left entirely to the model's own reasoning, without an external signal to anchor or correct it.
The early exit data make this concrete. On four of the six instances where the agentic approach halts early judging the problem unsolvable, the direct approach finds a valid plan, demonstrating that the self-assessment is factually incorrect.

Prior work indicates that self-correction without external verification is unreliable on planning tasks.
Iterative self-refinement with self-generated feedback can improve language generation outputs~\cite{madaan2023self}, but gains on planning tasks trace primarily to repeated sampling under an external verifier rather than to the self-critique itself, and iterative refinement of unexecuted plans degrades quality as critique rounds increase~\cite{stechly2024self}.
Stechly et al.\ evaluate self-refinement of \emph{unexecuted} plan text; the agentic approach here executes actions and observes real state transitions, a mechanistically distinct process~\cite{stechly2024self}.
Both cases nonetheless share the same root limitation: the absence of an external grounding signal.
The modest three percentage point gain observed here is consistent with this account. The interactive loop provides an opportunity for self-reflection, but without external grounding that reflection is insufficient to recover from planning errors.
This suggests that the large gains of coding agents stem from the objectivity of environmental feedback rather than from superior reasoning, and that step-wise agentic interaction alone, without a strong external signal, is not sufficient to substantially improve planning performance.

\subsection{Implications for Robotic Deployment}

A compelling near-term vision for autonomous robotics positions the LLM as the deliberative layer of a robotic system, equipped with tools and perception interfaces, applying broad world knowledge to coordinate specialised capabilities toward open-ended goals.
This vision implicitly assumes that the LLM's broad world knowledge, combined with step-wise environmental interaction, is sufficient for the model to assess whether its current trajectory is productive.
Our results challenge that assumption. Broad world knowledge does not substitute for externally grounded feedback, and an LLM cannot reliably determine from state observations alone whether it is making progress toward a goal.
The modest agentic gains observed here, concentrated in specific difficulty regimes and absent on others, confirm that step-wise environmental interaction alone is insufficient when that interaction provides no external signal about progress or failure.
The critical factor is not interaction per se, but whether the feedback the agent receives is externally grounded.
In robotic systems, this places a concrete design requirement on the perception layer. 
Sensors and state estimation must not merely report the current world state, but produce signals that tell the LLM whether it is making progress toward the goal or has entered a failing trajectory.
The performance ceiling of an LLM planning agent in a robotic system is therefore set not by the model's world knowledge, but by the richness and interpretability of the feedback the surrounding system provides.

\subsection{Limitations and Open Questions}

We test a single model tier (Claude Haiku 4.5) and results may differ for larger or more capable models~\cite{valmeekam2024llms}.
Each instance is evaluated in a single run and stochastic variance across runs is not measured.
The agentic approach exits early on six instances (32, 88, 89, 96, 98, 100), and this behaviour may not reproduce across runs.
Two directions for follow-up are directly motivated by these gaps:
\begin{itemize}
  \item Augmenting PyPDDLEngine with goal-distance heuristics in the per-step feedback would provide a direct test of whether externally grounded progress signals improve agentic success rates.
  \item Evaluating LLM planning across domains varying in training-data coverage would test the memorisation account more rigorously.
\end{itemize}

%% file: sections/6_conclusion.tex
\section{Conclusion}

We evaluated four approaches on 102 IPC Blocksworld instances under a 180-second budget to assess whether agentic step-wise interaction improves LLM planning over direct generation.
Fast Downward achieved 85.3\% success. The direct and agentic LLM approaches (Claude Haiku 4.5) achieved 63.7\% and 66.7\%, a modest advantage concentrated in the mid-difficulty range at $5.7\times$ higher token cost per solution.
Across most co-solved difficulty blocks, both LLM approaches produced shorter plans than seq-sat-lama-2011 despite its iterative plan quality improvement.
We release PyPDDLEngine as open-source software to make this class of experiment reproducible and extendable, in particular for testing whether richer per-step feedback changes the result.

The agentic advantage is not uniform. It concentrates in a specific difficulty regime, is absent on easy instances, and is selective on the hardest instances, producing unique solutions on three extreme cases (instances 76, 78, and 101) but failing on the majority.
The contrast between the large gains of coding agents and the modest gains here suggests that the value of agentic interaction scales with the quality and directionality of environmental feedback, which is externally grounded and objective in software execution but self-generated and unverified in PDDL step interaction.
The plan quality results, where both LLM approaches produce shorter plans than seq-sat-lama-2011 across most co-solved difficulty blocks despite having no quality-improvement mechanism, are consistent with the memorisation account and suggest that LLM performance on Blocksworld may reflect training-data recall rather than generalisable planning.
Taken together, current LLM planning agents appear to function as adaptive navigators of familiar problem spaces rather than general-purpose planners~\cite{kambhampati2024llms}.
Distinguishing genuine planning competence from training-distribution recall remains an open and important challenge for the field~\cite{valmeekam2023planning,valmeekam2024llms}.

%% file: sections/mybib.bib
@article{helmert2006fast,
  author    = {Helmert, Malte},
  title     = {The Fast Downward Planning System},
  journal   = {Journal of Artificial Intelligence Research},
  volume    = {26},
  pages     = {191--246},
  year      = {2006}
}

@misc{babu2025adaptive,
  author        = {Babu, Harisankar and Schillinger, Philipp and Asfour, Tamim},
  title         = {Adaptive Domain Modeling with Language Models: {A} Multi-Agent Approach to Task Planning},
  year          = {2025},
  eprint        = {2506.19592},
  archivePrefix = {arXiv},
  primaryClass  = {cs.AI}
}

@misc{lamalfa2025endtoend,
  author        = {La Malfa, Emanuele and Zhu, Ping and Marro, Samuele and Bernardini, Sara and Wooldridge, Michael},
  title         = {An End-to-end Planning Framework with Agentic {LLMs} and {PDDL}},
  year          = {2025},
  eprint        = {2512.09629},
  archivePrefix = {arXiv},
  primaryClass  = {cs.AI}
}

@misc{liu2024delta,
  author        = {Liu, Yuchen and Palmieri, Luigi and Koch, Sebastian and Georgievski, Ilche and Aiello, Marco},
  title         = {{DELTA}: Decomposed Efficient Long-Term Robot Task Planning using Large Language Models},
  year          = {2024},
  eprint        = {2404.03275},
  archivePrefix = {arXiv},
  primaryClass  = {cs.RO}
}

@misc{smirnov2024generating,
  author        = {Smirnov, Pavel and Joublin, Frank and Ceravola, Antonello and Gienger, Michael},
  title         = {Generating Consistent {PDDL} Domains with Large Language Models},
  year          = {2024},
  eprint        = {2404.07751},
  archivePrefix = {arXiv},
  primaryClass  = {cs.RO}
}

@misc{ding2023integrating,
  author        = {Ding, Yan and Zhang, Xiaohan and Amiri, Saeid and Cao, Nieqing and Yang, Hao and Kaminski, Andy and Esselink, Chad and Zhang, Shiqi},
  title         = {Integrating Action Knowledge and {LLMs} for Task Planning and Situation Handling in Open Worlds},
  year          = {2023},
  eprint        = {2305.17590},
  archivePrefix = {arXiv},
  primaryClass  = {cs.RO}
}

@inproceedings{mahdavi2024leveraging,
  author        = {Mahdavi, Sadegh and Aoki, Raquel and Tang, Keyi and Cao, Yanshuai},
  title         = {Leveraging Environment Interaction for Automated {PDDL} Translation and Planning with Large Language Models},
  booktitle     = {Advances in Neural Information Processing Systems},
  year          = {2024},
  eprint        = {2407.12979},
  archivePrefix = {arXiv},
  primaryClass  = {cs.LG}
}

@inproceedings{guan2023leveraging,
  author        = {Guan, Lin and Valmeekam, Karthik and Sreedharan, Sarath and Kambhampati, Subbarao},
  title         = {Leveraging Pre-trained Large Language Models to Construct and Utilize World Models for Model-based Task Planning},
  booktitle     = {Advances in Neural Information Processing Systems},
  year          = {2023},
  eprint        = {2305.14909},
  archivePrefix = {arXiv},
  primaryClass  = {cs.AI}
}

@misc{chen2023autotamp,
  author        = {Chen, Yongchao and Arkin, Jacob and Dawson, Charles and Zhang, Yang and Roy, Nicholas and Fan, Chuchu},
  title         = {{AutoTAMP}: Autoregressive Task and Motion Planning with {LLMs} as Translators and Checkers},
  year          = {2023},
  eprint        = {2306.06531},
  archivePrefix = {arXiv},
  primaryClass  = {cs.RO}
}

@inproceedings{ishay2025llmal,
  author        = {Ishay, Adam and Lee, Joohyung},
  title         = {{LLM+AL}: Bridging Large Language Models and Action Languages for Complex Reasoning about Actions},
  booktitle     = {Proceedings of the AAAI Conference on Artificial Intelligence},
  year          = {2025},
  eprint        = {2501.00830},
  archivePrefix = {arXiv},
  primaryClass  = {cs.CL}
}

@misc{chu2025llmmap,
  author        = {Chu, Kun and Zhao, Xufeng and Weber, Cornelius and Wermter, Stefan},
  title         = {{LLM+MAP}: Bimanual Robot Task Planning using Large Language Models and Planning Domain Definition Language},
  year          = {2025},
  eprint        = {2503.17309},
  archivePrefix = {arXiv},
  primaryClass  = {cs.RO}
}

@misc{huang2024planning,
  author        = {Huang, Sukai and Lipovetzky, Nir and Cohn, Trevor},
  title         = {Planning in the Dark: {LLM}-Symbolic Planning Pipeline without Experts},
  year          = {2024},
  eprint        = {2409.15915},
  archivePrefix = {arXiv},
  primaryClass  = {cs.AI}
}

@inproceedings{hao2024planning,
  author        = {Hao, Yilun and Zhang, Yang and Fan, Chuchu},
  title         = {Planning Anything with Rigor: General-Purpose Zero-Shot Planning with {LLM}-based Formalized Programming},
  booktitle     = {International Conference on Learning Representations},
  year          = {2025},
  eprint        = {2410.12112},
  archivePrefix = {arXiv},
  primaryClass  = {cs.AI}
}

@misc{singh2022progprompt,
  author        = {Singh, Ishika and Blukis, Valts and Mousavian, Arsalan and Goyal, Ankit and Xu, Danfei and Tremblay, Jonathan and Fox, Dieter and Thomason, Jesse and Garg, Animesh},
  title         = {{ProgPrompt}: Generating Situated Robot Task Plans using Large Language Models},
  year          = {2022},
  eprint        = {2209.11302},
  archivePrefix = {arXiv},
  primaryClass  = {cs.RO}
}

@misc{kwon2024fast,
  author        = {Kwon, Minseo and Kim, Yaesol and Kim, Young J.},
  title         = {Fast and Accurate Task Planning using Neuro-Symbolic Language Models and Multi-level Goal Decomposition},
  year          = {2024},
  eprint        = {2409.19250},
  archivePrefix = {arXiv},
  primaryClass  = {cs.RO}
}

@misc{tantakoun2025llms,
  author        = {Tantakoun, Marcus and Muise, Christian and Zhu, Xiaodan},
  title         = {{LLMs} as Planning Formalizers: {A} Survey for Leveraging Large Language Models to Construct Automated Planning Models},
  year          = {2025},
  eprint        = {2503.18971},
  archivePrefix = {arXiv},
  primaryClass  = {cs.AI}
}

@misc{valmeekam2023planning,
  author        = {Valmeekam, Karthik and Sreedharan, Sarath and Marquez, Matthew and Olmo, Alberto and Kambhampati, Subbarao},
  title         = {On the Planning Abilities of Large Language Models: {A} Critical Investigation with a Proposed Benchmark},
  year          = {2023},
  eprint        = {2302.06706},
  archivePrefix = {arXiv},
  primaryClass  = {cs.AI}
}

@misc{pallagani2024prospects,
  author        = {Pallagani, Vishal and Roy, Kaushik and Muppasani, Bharath and Fabiano, Francesco and Loreggia, Andrea and Murugesan, Keerthiram and Srivastava, Biplav and Rossi, Francesca and Horesh, Lior and Sheth, Amit},
  title         = {On the Prospects of Incorporating Large Language Models ({LLMs}) in Automated Planning and Scheduling ({APS})},
  year          = {2024},
  eprint        = {2401.02500},
  archivePrefix = {arXiv},
  primaryClass  = {cs.AI}
}

@inproceedings{correa2025classical,
  author        = {Corr{\^e}a, Augusto B. and Pereira, Andr{\'e} G. and Seipp, Jendrik},
  title         = {Classical Planning with {LLM}-Generated Heuristics: Challenging the State of the Art with {Python} Code},
  booktitle     = {Advances in Neural Information Processing Systems},
  year          = {2025},
  eprint        = {2503.18809},
  archivePrefix = {arXiv},
  primaryClass  = {cs.AI}
}

@inproceedings{silver2023generalized,
  author        = {Silver, Tom and Dan, Soham and Srinivas, Kavitha and Tenenbaum, Joshua B. and Kaelbling, Leslie Pack and Katz, Michael},
  title         = {Generalized Planning in {PDDL} Domains with Pretrained Large Language Models},
  booktitle     = {Proceedings of the AAAI Conference on Artificial Intelligence},
  year          = {2024},
  eprint        = {2305.11014},
  archivePrefix = {arXiv},
  primaryClass  = {cs.AI}
}

@misc{armony2025how,
  author        = {Armony, Ma'ayan and Mero{\~n}o-Pe{\~n}uela, Albert and Canal, Gerard},
  title         = {How Far Are {LLMs} from Symbolic Planners? {An} {NLP}-Based Perspective},
  year          = {2025},
  eprint        = {2508.01300},
  archivePrefix = {arXiv},
  primaryClass  = {cs.AI}
}

@misc{zhou2023isrllm,
  author        = {Zhou, Zhehua and Song, Jiayang and Yao, Kunpeng and Shu, Zhan and Ma, Lei},
  title         = {{ISR-LLM}: Iterative Self-Refined Large Language Model for Long-Horizon Sequential Task Planning},
  year          = {2023},
  eprint        = {2308.13724},
  archivePrefix = {arXiv},
  primaryClass  = {cs.RO}
}

@misc{liu2023llmp,
  author        = {Liu, Bo and Jiang, Yuqian and Zhang, Xiaohan and Liu, Qiang and Zhang, Shiqi and Biswas, Joydeep and Stone, Peter},
  title         = {{LLM+P}: Empowering Large Language Models with Optimal Planning Proficiency},
  year          = {2023},
  eprint        = {2304.11477},
  archivePrefix = {arXiv},
  primaryClass  = {cs.AI}
}

@misc{song2022llmplanner,
  author        = {Song, Chan Hee and Wu, Jiaman and Washington, Clayton and Sadler, Brian M. and Chao, Wei-Lun and Su, Yu},
  title         = {{LLM-Planner}: Few-Shot Grounded Planning for Embodied Agents with Large Language Models},
  year          = {2022},
  eprint        = {2212.04088},
  archivePrefix = {arXiv},
  primaryClass  = {cs.AI}
}

@inproceedings{zhao2023large,
  author        = {Zhao, Zirui and Lee, Wee Sun and Hsu, David},
  title         = {Large Language Models as Commonsense Knowledge for Large-Scale Task Planning},
  booktitle     = {Advances in Neural Information Processing Systems},
  year          = {2023},
  eprint        = {2305.14078},
  archivePrefix = {arXiv},
  primaryClass  = {cs.AI}
}

@inproceedings{kambhampati2024llms,
  author        = {Kambhampati, Subbarao and Valmeekam, Karthik and Guan, Lin and Verma, Mudit and Stechly, Kaya and Bhambri, Siddhant and Saldyt, Lucas and Murthy, Anil},
  title         = {Position: {LLMs} Can't Plan, But Can Help Planning in {LLM}-Modulo Frameworks},
  booktitle     = {Proceedings of the 41st International Conference on Machine Learning},
  series        = {Proceedings of Machine Learning Research},
  volume        = {235},
  year          = {2024},
  eprint        = {2402.01817},
  archivePrefix = {arXiv},
  primaryClass  = {cs.AI}
}

@misc{valmeekam2024llms,
  author        = {Valmeekam, Karthik and Stechly, Kaya and Kambhampati, Subbarao},
  title         = {{LLMs} Still Can't Plan; Can {LRMs}? {A} Preliminary Evaluation of {OpenAI}'s o1 on {PlanBench}},
  year          = {2024},
  eprint        = {2409.13373},
  archivePrefix = {arXiv},
  primaryClass  = {cs.AI}
}

@misc{chen2023llmstate,
  author        = {Chen, Siwei and Xiao, Anxing and Hsu, David},
  title         = {{LLM-State}: Open World State Representation for Long-horizon Task Planning with Large Language Model},
  year          = {2023},
  eprint        = {2311.17406},
  archivePrefix = {arXiv},
  primaryClass  = {cs.RO}
}

@misc{stechly2024self,
  author        = {Stechly, Kaya and Valmeekam, Karthik and Kambhampati, Subbarao},
  title         = {On the Self-Verification Limitations of Large Language Models on Reasoning and Planning Tasks},
  year          = {2024},
  eprint        = {2402.08115},
  archivePrefix = {arXiv},
  primaryClass  = {cs.AI}
}

@misc{li2024lasp,
  author        = {Li, Haoming and Chen, Zhaoliang and Zhang, Jonathan and Liu, Fei},
  title         = {{LASP}: Surveying the State-of-the-Art in Large Language Model-Assisted {AI} Planning},
  year          = {2024},
  eprint        = {2409.01806},
  archivePrefix = {arXiv},
  primaryClass  = {cs.AI}
}

@misc{wei2025plangenllms,
  author        = {Wei, Hui and Zhang, Zihao and He, Shenghua and Xia, Tian and Pan, Shijia and Liu, Fei},
  title         = {{PlanGenLLMs}: A Modern Survey of {LLM} Planning Capabilities},
  year          = {2025},
  eprint        = {2502.11221},
  archivePrefix = {arXiv},
  primaryClass  = {cs.AI}
}

@article{madaan2023self,
  author        = {Madaan, Aman and Tandon, Niket and Gupta, Prakhar and Hallinan, Skyler and Gao, Luyu and Wiegreffe, Sarah and Alon, Uri and Dziri, Nouha and Prabhumoye, Shrimai and Yang, Yiming and others},
  title         = {Self-Refine: Iterative Refinement with Self-Feedback},
  journal       = {Advances in Neural Information Processing Systems},
  volume        = {36},
  pages         = {46534--46594},
  year          = {2023}
}

@inproceedings{yao2022react,
  author        = {Yao, Shunyu and Zhao, Jeffrey and Yu, Dian and Du, Nan and Shafran, Izhak and Narasimhan, Karthik and Cao, Yuan},
  title         = {{ReAct}: Synergizing Reasoning and Acting in Language Models},
  booktitle     = {International Conference on Learning Representations},
  year          = {2023},
  eprint        = {2210.03629},
  archivePrefix = {arXiv},
  primaryClass  = {cs.CL}
}

@inproceedings{shinn2023reflexion,
  author        = {Shinn, Noah and Cassano, Federico and Gopinath, Ashwin and Narasimhan, Karthik and Yao, Shunyu},
  title         = {Reflexion: Language Agents with Verbal Reinforcement Learning},
  booktitle     = {Advances in Neural Information Processing Systems},
  year          = {2023},
  eprint        = {2303.11366},
  archivePrefix = {arXiv},
  primaryClass  = {cs.AI}
}

@inproceedings{yang2024sweagent,
  author        = {Yang, John and Jimenez, Carlos E. and Wettig, Alexander and Lieret, Kilian and Yao, Shunyu and Narasimhan, Karthik and Press, Ofir},
  title         = {{SWE}-agent: Agent-Computer Interfaces Enable Automated Software Engineering},
  booktitle     = {Advances in Neural Information Processing Systems},
  year          = {2024},
  eprint        = {2405.15793},
  archivePrefix = {arXiv},
  primaryClass  = {cs.SE}
}

@misc{wang2023voyager,
  author        = {Wang, Guanzhi and Xie, Yuqi and Jiang, Yunfan and Mandlekar, Ajay and Xiao, Chaowei and Zhu, Yuke and Fan, Linxi and Anandkumar, Anima},
  title         = {Voyager: An Open-Ended Embodied Agent with Large Language Models},
  year          = {2023},
  eprint        = {2305.16291},
  archivePrefix = {arXiv},
  primaryClass  = {cs.AI}
}
